\documentclass[letterpaper, 10 pt, conference]{ieeeconf}  

\IEEEoverridecommandlockouts                              

\usepackage{graphics} 
\usepackage{graphicx}
\usepackage{epsfig} 
\usepackage{mathptmx} 
\usepackage{times} 
\usepackage{amsmath} 
\usepackage{amssymb}  
\usepackage{xcolor}
\usepackage{multirow}
\usepackage{subfiles}
\usepackage{float}
\usepackage{cuted}
\usepackage{capt-of}
\usepackage{arydshln}

\newcommand{\modelfullname}{Grasping-and-Unlocking Model}
\newcommand{\modelname}{GUM}

\title{\LARGE \bf
DoorBot: Closed-Loop Task Planning and Manipulation for \\ Door Opening in the Wild with Haptic Feedback}

\author{
    Zhi Wang$^{1,2*}$\thanks{$^1$~Zhi Wang, Yuchen Mo, and Shengmiao Jin are with University of Illinois Urbana-Champaign \tt\small \{zhi, yuchenm7, jin45, yuanwz\}@illinois.edu},
    Yuchen Mo$^{1*}$\thanks{$^2$~Zhi Wang is also with Tsinghua Univeristy \tt\small tx.leo.wz@gmail.com},
    Shengmiao Jin$^{1}$,
    Wenzhen Yuan$^{1}$\thanks{$^*$~Equal Contribution.}
}

\begin{document}
\maketitle

\begin{strip}
\begin{minipage}{\textwidth}\centering
\vspace{-50pt}
\includegraphics[width=0.85\textwidth]{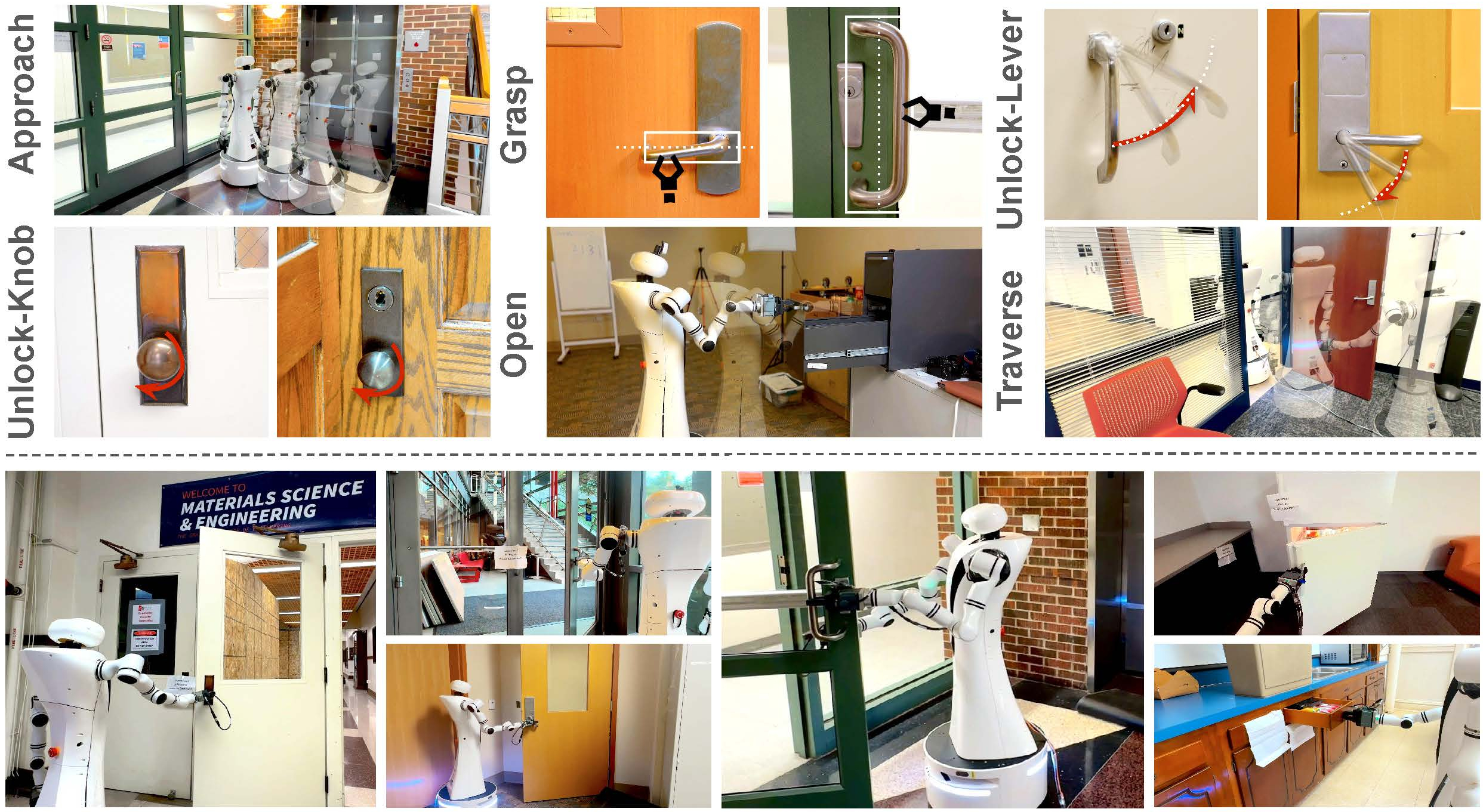}
\captionof{figure}{
We propose a hierarchical closed-loop controller to help a mobile robot automatically open various doors and walk through them in open environments. Our method can robustly generalize to different handles in the wild.
}
\label{fig:teaser}
\end{minipage}
\end{strip}

\thispagestyle{empty}
\pagestyle{empty}

\begin{abstract}

Robots operating in unstructured environments face significant challenges when interacting with everyday objects like doors. 
They particularly struggle to generalize across diverse door types and conditions.
Existing vision-based and open-loop planning methods often lack the robustness to handle varying door designs, mechanisms, and push/pull configurations. 
In this work, we propose a haptic-aware closed-loop hierarchical control framework that enables robots to explore and open different unseen doors in the wild. 
Our approach leverages real-time haptic feedback, allowing the robot to adjust its strategy dynamically based on force feedback during manipulation. 
We test our system on 20 unseen doors across different buildings, featuring diverse appearances and mechanical types. 
Our framework achieves a 90\% success rate, demonstrating its ability to generalize and robustly handle varied door-opening tasks. 
This scalable solution offers potential applications in broader open-world articulated object manipulation tasks.

\end{abstract}

\section{Introduction}
\label{sec:intro}
Deploying robot systems in open environments has long been a key challenge in robotics, requiring solutions for unstructured settings and varied object types. 
Generalizing to complex tasks like door opening is particularly difficult due to the diverse designs, mechanisms, and conditions of articulated objects.
Researchers have explored this through kinematics-based methods that depend on known door models~\cite{6385835,abraham2020model,sturm2011probabilistic}, as well as end-to-end learning methods that generalize to visually similar doors~\cite{kang2024versatile,gu2017deep,li2020hrl4in,chebotar2017path}. More recently, large language models (LLMs) and vision-language models (VLMs) have been used for high-level planning, enabling robots to execute long-horizon tasks like door-opening in new environments~\cite{ahn2022can,liang2023code,driess2023palm,brohan2023rt}.

Despite successes, these approaches are often limited by their reliance on visual data and predefined models, making it difficult to adapt to diverse and unstructured settings. 
Robots still struggle to address non-visual properties, such as internal mechanisms or unexpected resistance. 
For example, a robot might need to discern when to pull rather than push or when to rotate a handle in an unconventional direction—tasks that require immediate adaptation to unanticipated scenarios.

While current methods, especially those reliant on open-loop control, face difficulties generalizing to these scenarios, humans effortlessly solve them using an explore-and-adapt strategy based on haptic feedback. 
They adjust actions like switching between pushing and pulling, and achieve nearly 100\% success rates. 
Inspired by this capability, we pose the question: \textbf{Can robots similarly learn to explore and adapt to manipulation tasks using haptic feedback?}

To address this, we propose a haptic-aware closed-loop control framework that combines vision-based perception with real-time haptic feedback. 
Our haptic perception module, using joint current readings and gripper resistance, dynamically adapts to non-visual properties like unexpected resistance, ensuring robust performance where visual data alone may be insufficient. 
Our vision module, trained on a small and easily accessible dataset, generalizes effectively to unseen door types with minimal adaptation, reducing the need for large-scale real robot data collection. 
This combination enables accurate grasp pose predictions and enhances performance in unstructured environments.

Field tests on 20 unseen doors across a university campus showed a 40\% improvement in success rate over baseline methods, demonstrating the effectiveness of integrating haptic feedback into a closed-loop control system. 
These results highlight the power of combining vision-based perception and real-time haptic feedback to substantially enhance a robot’s ability to generalize and adapt in complex, real-world environments.
In the long term, this work contributes to more adaptable, general-purpose robots capable of performing various tasks in unpredictable settings. 
The combination of vision-based perception and haptic feedback offers a path toward more autonomous robots, capable of operating independently in real-world scenarios that require both generalizable perception and dynamic adaptability.

\section{Related works}
\label{sec:relatedworks}
\subsection{Door manipulation systems}

Manipulating articulated objects, such as doors and drawers, is a fundamental skill for robots in domestic environments \cite{arduengo2021robust,stuede2019door,5649847}. 
Various methods have been proposed for door-opening tasks. 
Kinematic-based approaches \cite{6385835,abraham2020model,sturm2011probabilistic} assume knowledge of the door’s model or use online system identification, while geometric-based methods \cite{tremblay2018deep,chu2019learning} extract 3D pose information to generate trajectories. 
Although effective with accurate priors, these methods struggle to generalize across different shapes and environments. 
Keypoint-based approaches \cite{9206039} mitigate this but face data collection challenges due to their reliance on RGB-D data.

End-to-end imitation learning (IL) and reinforcement learning (RL) \cite{kang2024versatile,gu2017deep,li2020hrl4in,chebotar2017path} have also been applied to manipulation tasks, though RL-based methods often face issues when transferring from simulation to real-world tasks\cite{qin2023dexpoint, urakami2019doorgym}. 
Newer pipelines combine IL and RL to fine-tune policies on real robots \cite{xiong2024adaptive}. 
More recently, LLMs and VLMs have been applied to long-horizon manipulation tasks \cite{ahn2022can,liang2023code,driess2023palm,brohan2023rt}, with LLMs serving as high-level planners and VLMs handling both visual perception and planning.
We will compare our performance to a large-model baseline in Section \ref{sec:experiments}.

Our approach integrates real-time haptic feedback into a closed-loop control framework, allowing robot dynamic adaptation to non-visual properties. 
In contrast to methods requiring large amounts of data or pre-defined models, our system uses vision models trained primarily on a small set of Internet RGB images, which generalizes effectively to real-world doors. 
By leveraging Dynamic Movement Primitives (DMPs), we further reduce the need for expert demonstrations, improving the system's adaptability in unstructured environments.

\subsection{Haptic feedback control}

Humans rely heavily on haptic and tactile feedback when manipulating objects, allowing them to perform some contact-rich tasks without visual guidance. 
Inspired by this, the robotics community has explored haptic and tactile sensors and feedback systems for decades \cite{ding2021sim,yuan2017gelsight,calandra2018more}, enhancing robotic manipulation by providing information about contact forces, geometry, and textures.

For articulated object manipulation, particularly door opening, \cite{karayiannidis2016adaptive} uses adaptive velocity control based on force/torque feedback to manage uncertain kinematics. 
Similarly, \cite{jain2008behaviors} employs force/torque sensors to detect door states or classify door types, compensating for vision-based limitations. 
Other work \cite{calandra2018more} combines tactile and visual data to improve decision-making. 
Recent efforts, such as \cite{van2015learning} and \cite{xu2022towards}, incorporate haptic feedback into reinforcement learning (RL) to enhance manipulation policies.

Despite their effectiveness, tactile sensors and force/torque systems are expensive, difficult to implement, and often too fragile for tasks like door opening. 
In contrast, our closed-loop system leverages low-cost motor current data for haptic feedback, providing a practical and adaptable solution for various door types and conditions.


\section{Methods}
\label{sec:methods}

We propose a hierarchical control framework for the door-opening task.
This framework consists of a high-level feedback planner that coordinates the robot’s actions with six low-level motion primitives. 
Some of these primitives are supported by vision and haptic perception modules to form a closed-loop control system, allowing the system to perform real-time exploration and adaptation to the environment. 

\begin{figure*}[t]
\vspace{2mm}
    \centering
    \includegraphics[width=0.9\linewidth]{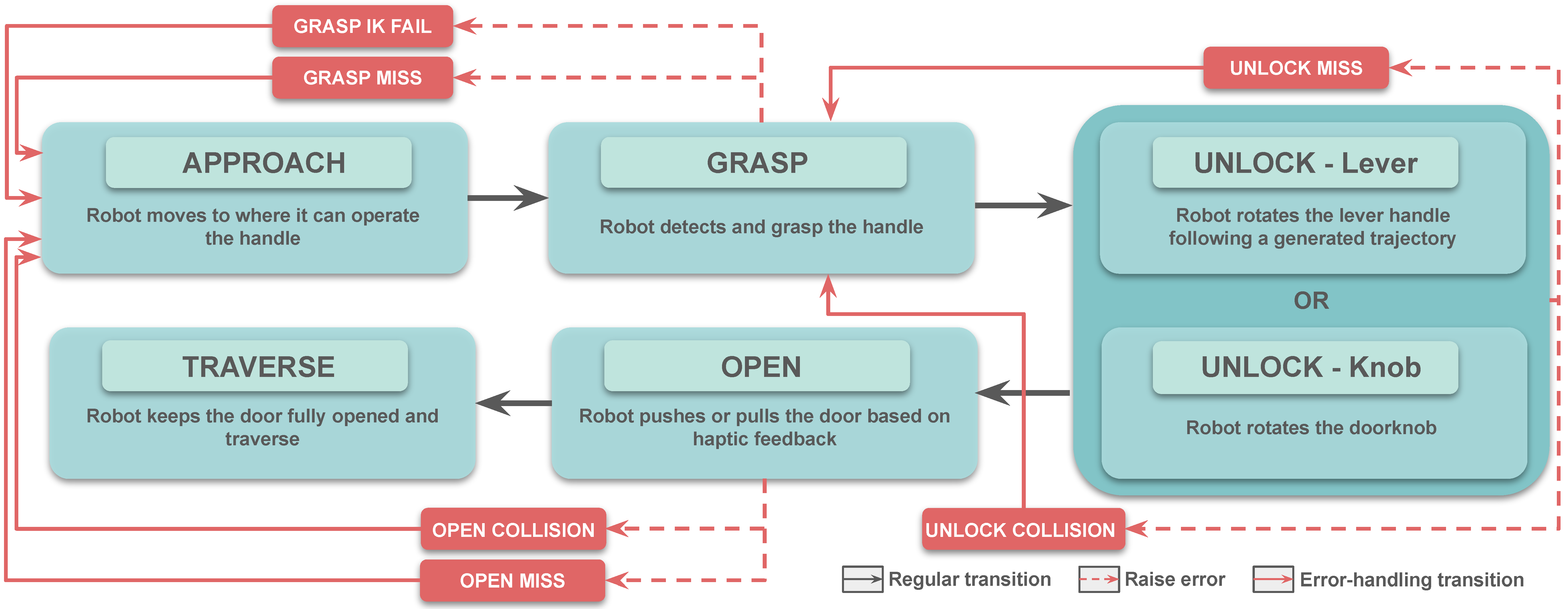}
    \caption{We design a state machine as the high-level planner. Aside from regular working transitions, we define error-handling transitions to recover from unexpected scenarios. For example, when \textit{OPEN\_COLLISION} is triggered due to collision in the \textit{open} primitive, the robot then transitions back to \textit{grasp}.}
    \label{fig:method_state_machine}
\end{figure*}

\subsection{High-level controller}

Our high-level controller plans the sequence of motion primitives, which are executed through the low-level controller. 
The high-level controller considers the feedback from the low-level controller, which might include the operation result and the key parameters.
We employ a state machine as a controllable, interpretable, and data-efficient solution for high-level planning, as illustrated in Fig. \ref{fig:method_state_machine}.
The state machine defines possible states and transitions, clustered by the corresponding motion primitives. 
Compared to previous works utilizing learning-based high-level policies \cite{xiong2024adaptive}, our method is much more data efficient. Moreover, the transitions incorporate feedback from the low-level controllers, allowing for more effective error handling during experiments.

\subsection{Low-level controller}
\label{subsec:motion_primitive}

We design six motion primitives based on the key steps of opening doors and implement them through low-level controllers. This reduces the dimensionality of the action space and avoids reliance on extensive human expert data.

\subsubsection{Approach}

This primitive navigates the robot to move towards the target door until it reaches a proper position for operating the door handle.
The robot starts with detecting the door and fits a plane of the door surface using RANSAC~\cite{fischler1981random}.
The result provides information on the door's 3D location and the distance to the door. 
The robot then moves to a preset position relative to the door.
This primitive does not trigger error feedback.

\subsubsection{Grasp}

We define \textit{grasp} as the primitive of grasping door handles, consisting of the detection of a proper grasp point and performing the grasp motion. 

To detect the grasp point, we first want to locate the door handle. We use pre-trained vision models, Detic~\cite{zhou2022detecting} and Segment Anything~\cite{kirillov2023segment} (SAM), to provide the type of handle and a mask of the handle area on the image.
Based on the handle type and shape, we need to refine the grasp point on the handle further, which should be a good position to operate the handle. 
We introduce the \modelfullname~(\modelname) model to fine-tune the grasp point, which is detailed in Sec. \ref{subsec:gum}.
The grasp orientation is determined based on the mask geometry and normal vector, followed by the execution of DMPs to reach the pose while avoiding potential collisions.

During the execution, the system may encounter two errors: \textit{GRASP\_IK\_FAIL} occurs if inverse kinematics fail for the predicted pose.
\textit{GRASP\_MISS} occurs if the robot fails to grasp the handle, detected by low resistance in the gripper.
Both errors trigger a state machine transition back to the \textit{approach} primitive for re-execution.

\subsubsection{Unlock - Lever}

This primitive allows the robot to unlock a lever-shaped handle by generating a circular trajectory based on the grasp pose, rotation axis, and radius predicted by \modelname~during the \textit{grasp} phase.
Joint motor current feedback is monitored throughout this process. 
A large current reading indicates a large impedance force from the handle, which is typically caused by the wrong direction of rotating or the door being locked. 
Then the robot tries to rotate the handle in the opposite direction.
Details of this mechanism are discussed in Sec. \ref{subsec:haptic}.

During this primitive, two error types may be triggered: 
\textit{UNLOCK\_MISS} occurs if the gripper loses contact with the handle. 
\textit{UNLOCK\_COLLISION} occurs if the gripper collides with the door or becomes obstructed, indicates by spiking joint motor current readings.

\subsubsection{Unlock - Knob}

This primitive is designed for unlocking doorknobs and follows a procedure similar to that of \textit{unlock-lever}, albeit with different constraints for trajectory generation.
Joint motor current feedback and a halting threshold ensure safe and efficient unlocking, with error feedback mechanisms similar to those in the \textit{unlock-lever} primitive.

\subsubsection{Open}

After successfully grasping and unlocking the door handle, the \textit{open} primitive targets at opening the door by pushing or pulling the handle in the gripper. The robot initiates a slight backward motion to determine whether the door is a push-type or pull-type based on haptic reading. 
The corresponding pre-defined mobile base movement is then executed to either push or pull the door open.

During this primitive, two errors may arise:
\textit{OPEN\_MISS} occurs if the gripper loses contact with the handle while opening, prompting a retry of the entire pipeline.
\textit{OPEN\_COLLISION} occurs if the gripper collides with the door, also leading to a transition back to the \textit{approach} primitive.

\subsubsection{Traverse}

This primitive enables the robot to pass through doors. 
For push-type doors, the robot simply moves forward to go through it.
For pull-type doors, a pre-defined sequential bimanual trajectory is executed after pulling the door to keep the door fully open, ensuring safe traversal.
No error feedback is associated with this primitive.

\subsection{\modelfullname}
\label{subsec:gum}

We propose the \modelfullname~(\modelname) to precisely detect the grasp point on the door handle and the potential motion trajectory to rotate the handle from the RGB input. 
This information is essential for the \textit{grasp} and two \textit{unlock} primitives. 

\begin{figure}[t]
\vspace{2mm}
\centering
\includegraphics[width=1.0\linewidth]{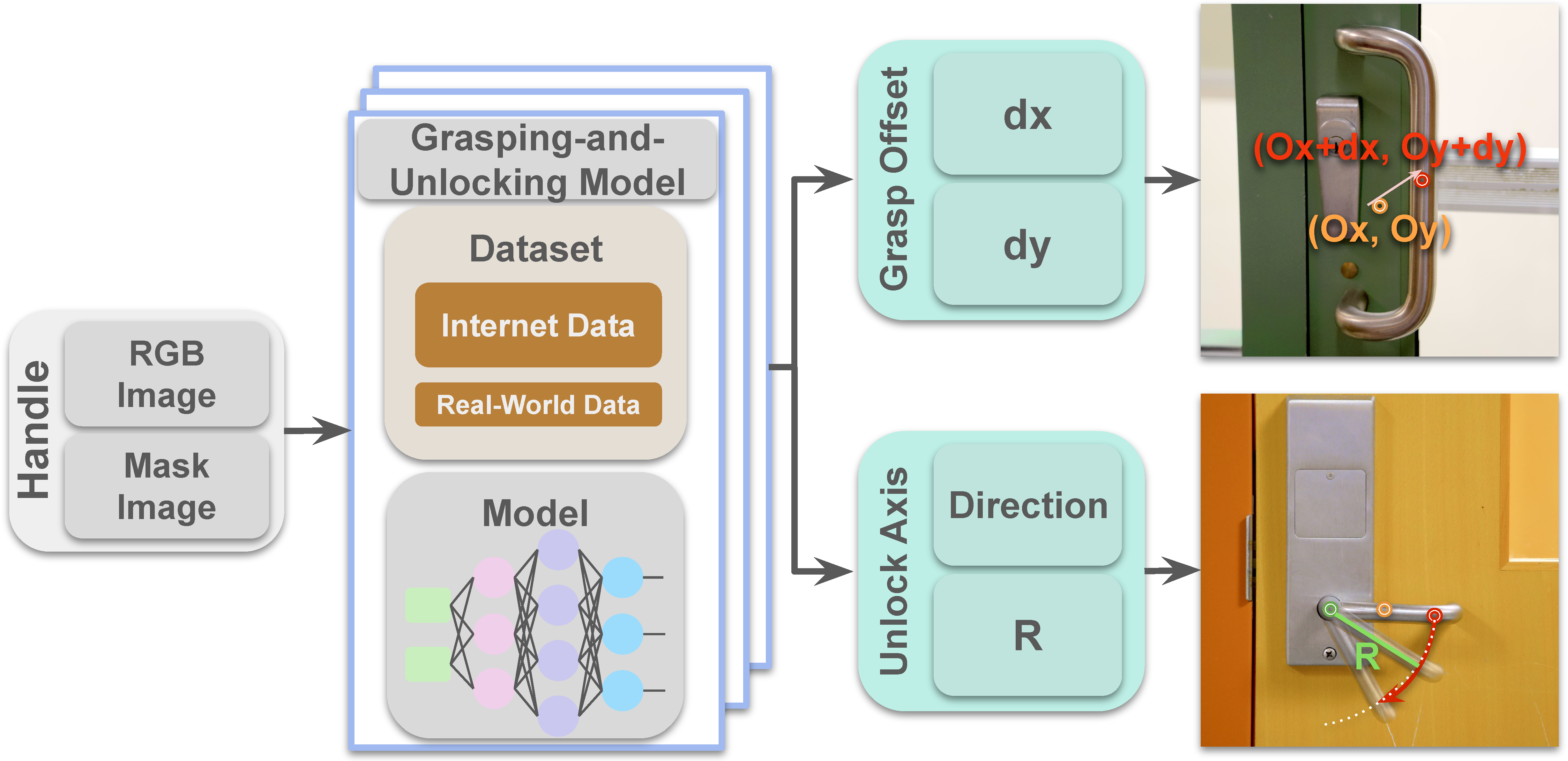}
\caption{Overview of \modelname. \modelname~refines the model-based grasp pose prior for the \textit{grasp} primitive, and simultaneously predicts the motion trajectory for unlocking the door handle.
}
\label{fig:gum_arch}
\vspace{-2mm}
\end{figure}

\modelname~takes the raw RGB image and object mask from Detic and SAM as input, utilizes a pre-trained ResNet-18 backbone to encode the inputs, and a 3-layer MLP to predict three key values:
the 2D offset from the mask centroid to the new grasp point $(\mathrm{d}x, \mathrm{d}y)$, and rotation parameter $R$.
The sign of $R$ indicates the unlocking direction (clockwise/counterclockwise), while the magnitude represents the radius of rotation. 
Fig. \ref{fig:gum_arch} shows the overall architecture.

To train \modelname, we create a dataset of 1,303 images featuring various door handles, collected from the Internet and real-world photos. 
The dataset includes four common handle types: lever handles, doorknobs, crossbars, and cabinet handles. 
Based on object masks generated by Detic and SAM, we manually label the appropriate grasp point and rotation parameters on the images. 
This data collection process ensures the model's generalizability by including a large collection of different handles, also avoids the time and labor cost of collecting data with a real robot.

\subsection{Haptic feedback}
\label{subsec:haptic}

\begin{figure}[t]
\vspace{2mm}
\centering
\includegraphics[width=0.8\linewidth]{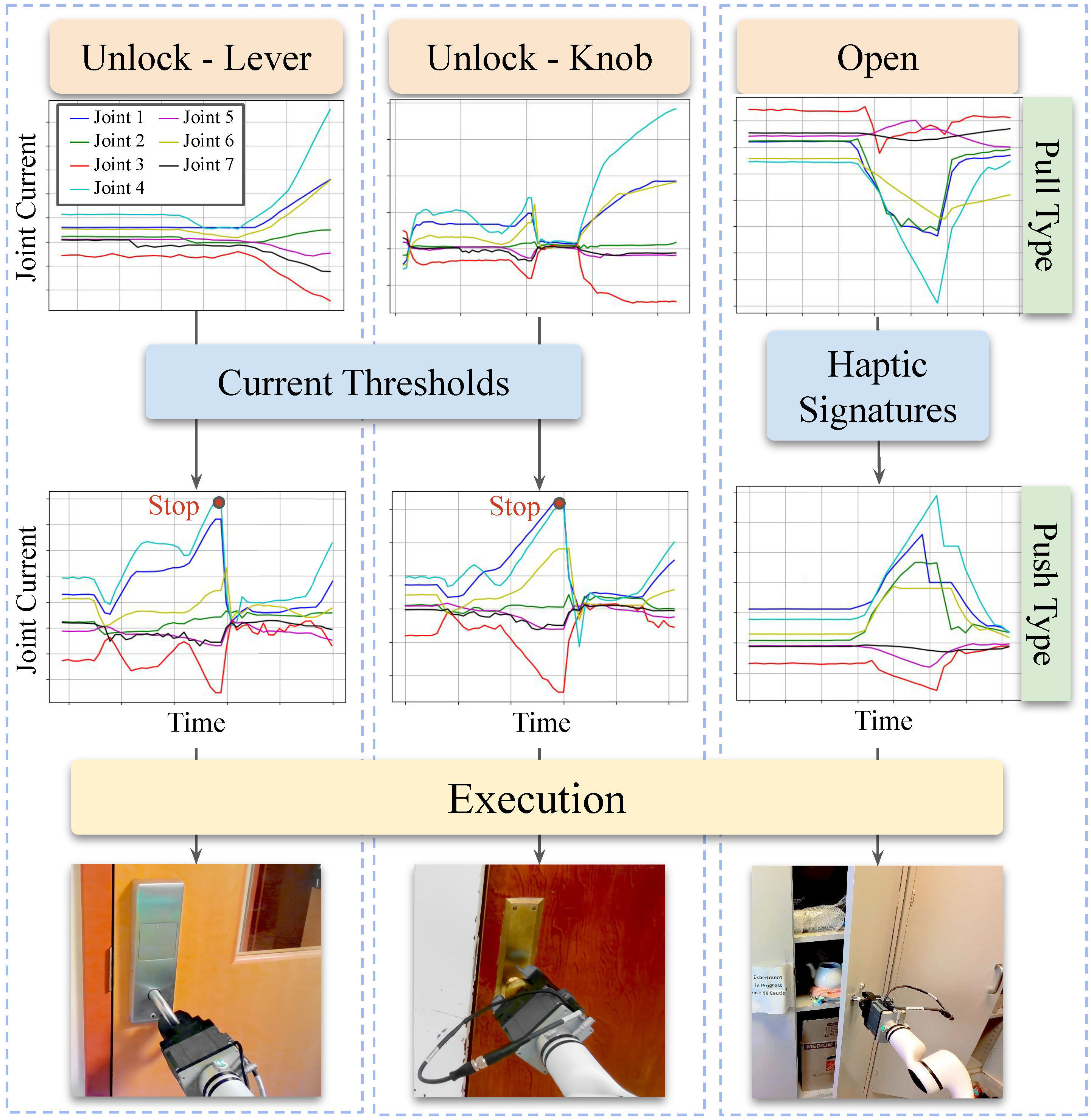}
\caption{Haptic feedback in 3 motion primitives. For \textit{unlock-lever} and \textit{unlock-knob}, the current threshold for the elbow joint tells the robot when to stop. For \textit{open} the increase/decrease of current feedback on the elbow joint shows the push-/pull-type of the door.}
\label{fig:method_haptic_feedback}
\end{figure}

As mentioned in Sec. \ref{subsec:motion_primitive}, we utilize haptic feedback for rotation termination, grasping state determination, and push/pull door classification.
Unlike prior work using specialized force/torque sensors \cite{jain2008behaviors}, we leverage cost-effective joint current feedback and gripper resistance feedback.

For example, when the robot attempts to rotate a handle that has reached its physical limit, the system, operating in position control mode, continues applying force to overcome the resistance, causing a spike in current readings, particularly in the elbow joint (joint 4). 
As shown in Fig. \ref{fig:method_haptic_feedback}, monitoring these current readings in real-time allows the system to reliably detect the appropriate termination of rotation. 
Preliminary tests established an appropriate current threshold to differentiate these cases from collisions.

Similarly, during the classification phase in the \textit{open} primitive, we distinguish between push-type and pull-type doors based on the distinct current signatures exerted on the joints. 
We conduct extensive experiments to demonstrate that this approach is both cost-efficient and effective for manipulating objects of varying weights and sizes.

\section{Experiments}
\label{sec:experiments}
In this section, we aim to answer the following questions:

\begin{itemize}
    \item Does the system generalize to various door types in real-world settings, given the limited size of the \modelname~training dataset? 
    \item How effective is the integration of haptic and other feedback modalities in the door-opening task? 
    \item Can VLMs effectively serve as high-level planners or robustly guide low-level motion?
\end{itemize}

\subsection{Experiment setting}

\begin{figure}[t]
    \centering
    \includegraphics[width=0.9\linewidth]{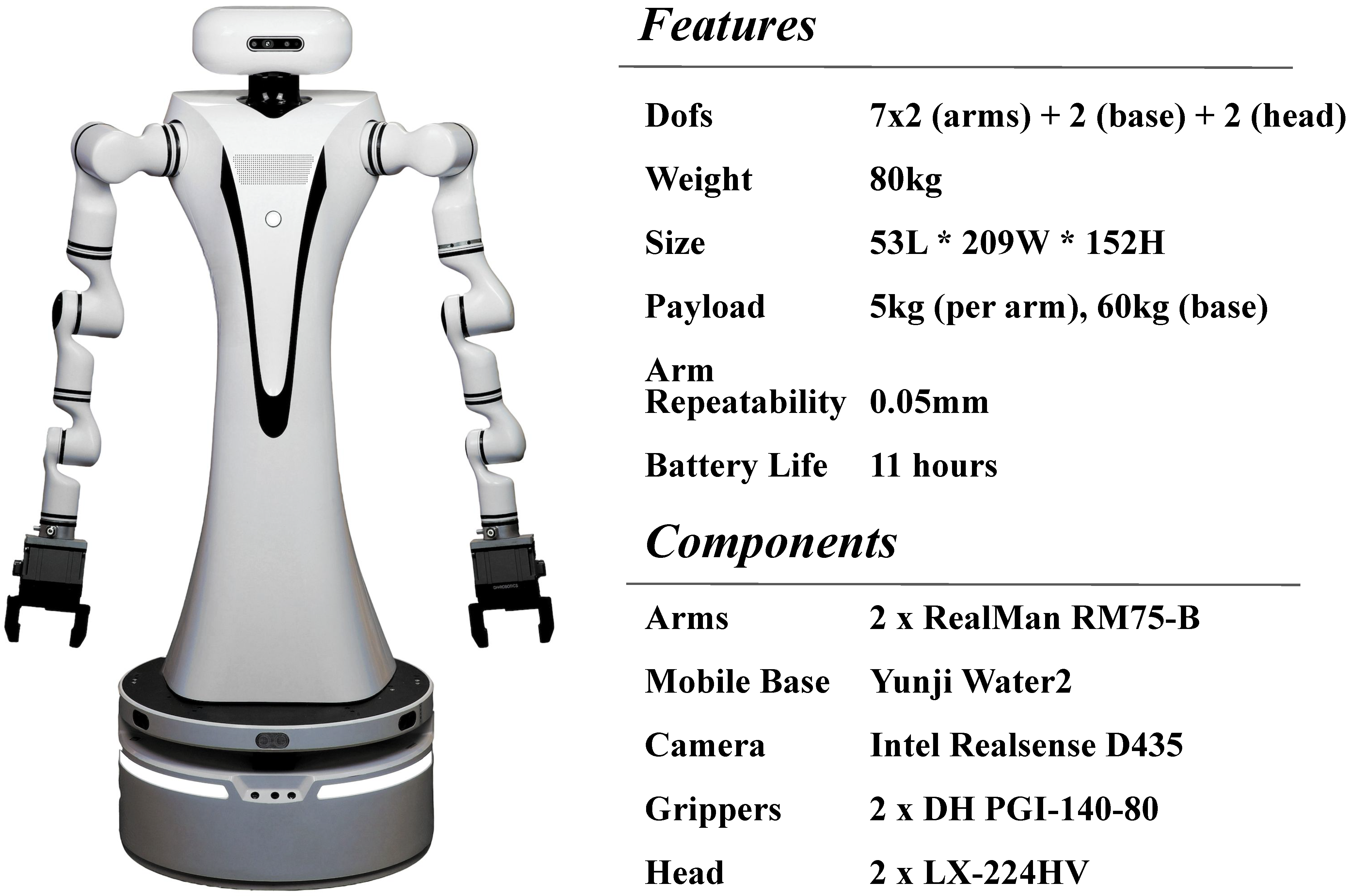}
    \caption{Visual appearance and configurations of our bimanual mobile robot. }
    \label{fig:system_appearance}
\end{figure}

We choose the RealMan dual-arm compound robot~\cite{RealMan2024CompoundRobot} as our hardware platform.
As shown in Fig. \ref{fig:system_appearance}, the robot is equipped with two 7-DoF arms, each with a 5kg payload, a mobile base, and a 2-DoF robot head with an Intel RealSense D435 camera.
Two DH PGI-140-80 grippers~\cite{DHRobotics2024PGI} are mounted on the arms to handle the door-opening tasks. 

\begin{figure}[t]
\vspace{2mm}
\centering
\includegraphics[width=1.0\linewidth]{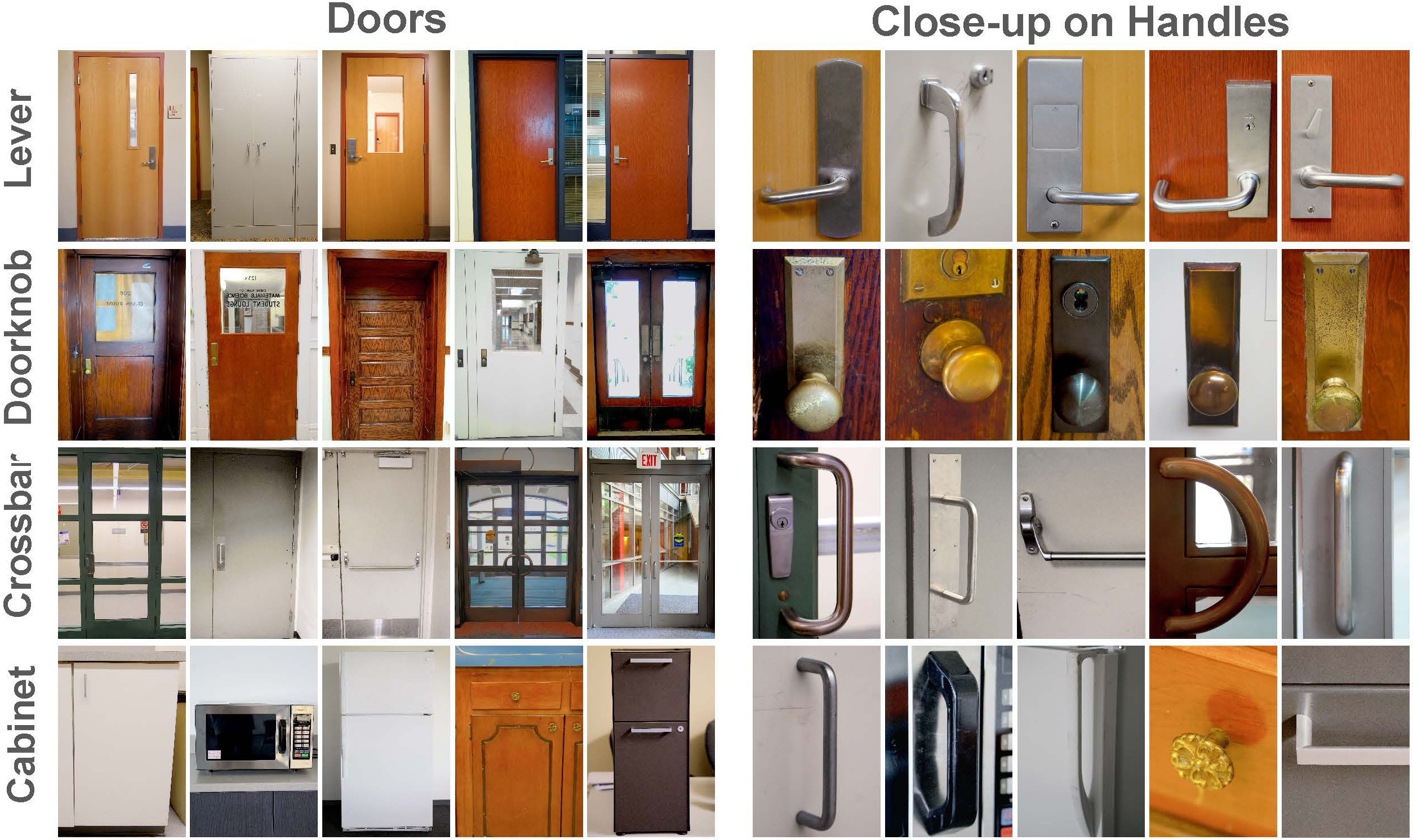}
\caption{Field test setting. We experimented with 20 environments on the university campus. These scenes in the wild contain various door appearances, handle types, physical properties, and visual distractions like illumination. None of these scenes have been seen in our training dataset.}
\label{fig:exp_setting}
\end{figure}

We conducted an extensive field study on a university campus to evaluate the system's efficacy. 
As shown in Fig. \ref{fig:exp_setting}, Our experiments encompassed 20 distinct doors distributed across 8 buildings. 
During each trial, the robot began from a random position near the door with an initial pose that ensures visibility of the door. 
The system operated autonomously until it either completed the task or performed a potentially hazardous action, 
defined as actions that could physically damage the robot or environment.
A trial is considered successful if it opens the door to an angle greater than 45 degrees.
Aside from doors, we also conducted experiments on some other articulated objects with handles like cabinets and drawers, as they share similar mechanisms.
For drawers, we consider the trial successful if it is fully opened.

\subsection{Baseline}

We employ Google Gemini 1.5 Pro~\cite{team2023gemini} as our baseline method. 
Previous works~\cite{ahn2022can,liang2023code,driess2023palm,brohan2023rt} have demonstrated that LLMs and VLMs can serve as high-level planners, coordinating low-level motion primitives without domain-specific expert data or extensive exploration.
In our setup, the VLM generates the next action based on prior actions and error feedback, acting as a high-level planner. 

Additionally, the VLM is used to predict grasp poses by providing it with visual inputs, text prompts, and example images, serving as an alternative to \modelname. 
As a low-level controller, we use the VLM in a few-shot in-context learning manner: feeding it text prompts, visual observations with masks, and task-specific input-output examples, and then extracting predictions from its responses.

\subsection{Overall success rate in field tests}

We evaluate the 4 different combinations of high- and low-level controllers by their success rate.
For each method, we test 5 trajectories per door, resulting in a total of 400 trajectories across all 20 doors.

\begin{table}[th]
\vspace{2mm}
\centering
\caption{Overall success rate}
\begingroup
\setlength{\tabcolsep}{5pt} 
\begin{tabular}{c|cccc|c}
Method & Crossbar & Lever & Doorknob & Cabinet & Avg                           \\ \hline
VLM+VLM                     & 28\%            & 28\%            & 52\%            & 92\%            & 50\%  \\
SM+VLM  & 32\%            & 52\%            & 76\%            & 92\%             & 63\%  \\
VLM+\modelname & 64\%            & 88\%            & \textbf{92\%}            & \textbf{100\%}            & 86\% \\  
\textbf{SM+\modelname~(Ours)} & \textbf{76\%}            & \textbf{92\%}            & \textbf{92\%}            & \textbf{100\%}            &\textbf{90\%} \\
\end{tabular}
\label{table:overall_success_rate}
\endgroup
\vspace{-3mm}
\end{table}

As summarized in Table \ref{table:overall_success_rate}, our method consistently outperforms other combinations, showing an average success rate improvement from 50\% to 90\% across all manipulation tasks. 
None of the doors nor the handles are seen in the training set, proving our model's generalizability across different situations. 
Moreover, in the rare case of a missed grasp, our state machine can often robustly recover from its failure, while other methods cannot.

Despite the success rate improvement to the baseline models, our model still fails in two specific cases:
a door with a
C-shaped crossbar where no method succeeds even once, and doors with large transparent areas, making its plane model hard to estimate with RGB-D camera. 
The first challenge arises from the inherent limitations of learning-based methods, as our dataset is not large enough to cover all shapes of handles, 
while detecting glass doors is a separate research problem beyond the scope of this work. 

In contrast, when using VLM as a few-shot alternative to our \modelname~model, its performance is notably lower.
On the other hand, when VLM is used as the high-level planner, it performs comparably to the state machine.
Despite these comparable results, the VLM exhibits several failure modes when used as a high-level planner, particularly in:
\begin{itemize}
    \item Lack of robustness: In some trials, VLM incorrectly predicts the \textit{approach} action immediately after \textit{grasp}, leading to task failures.
    \item Failure to integrate feedback: The VLM does not consistently act on feedback from the previous action, which results in inefficient or incorrect task execution.
\end{itemize}

Additionally, Table \ref{table:overall_success_rate} shows that crossbar and lever-handle types pose relatively greater challenges. 
To assess the impact of each module, we conduct modular tests and ablation experiments on a subset of 5 randomly sampled lever-handle and crossbar doors.

\subsection{Ablation: Evaluation of the \modelname~module}

To verify how much \modelname~help the robot generalize its grasp pose detection to various door handles, 
we compare the system's overall success rate under three conditions: with \modelname, with \modelname $^*$, and with full \modelname. The \modelname$^*$ model is trained with only Internet images. 
As shown in Table \ref{table:gum_result}, \modelname~drastically raised the success rate. 
It is also worth noting that the model trained solely on Internet images performed comparable to the full model, indicating that \modelname~can help manipulation in the open world with easy-to-get Internet RGB data.

\begin{figure}[t]
\vspace{2mm}
\centering
\includegraphics[width=1.0\linewidth]{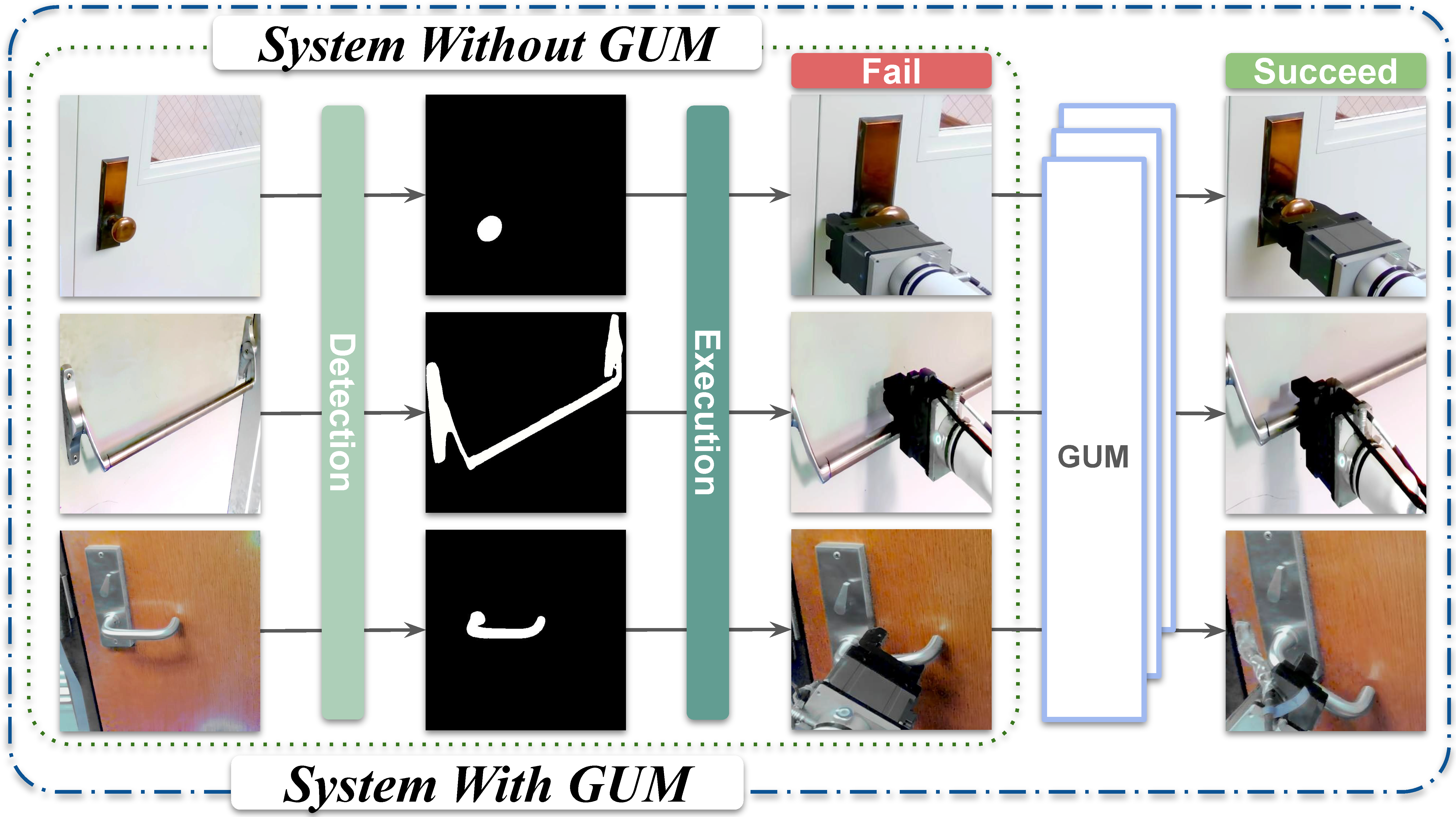}
\caption{Examples of how \modelname~fixes bad grasp pose during our field test. }
\label{fig:result_gum_effective}
\end{figure}

\begin{table}[ht]
\centering
\caption{Effectiveness of \modelname}
\begin{tabular}{c|ccccc|c}
                                      & \textbf{door1} & \textbf{door2} & \textbf{door3} & \textbf{door4} & \textbf{door5} & \textbf{sum}                           \\ \hline
w/o \modelname                     & 0/5            & 2/5            & 3/5            & 0/5            & 0/5            & 20\%  \\
\modelname$^*$  & 4/5            & 5/5            & 5/5            & 5/5            & 5/5            & 92\%  \\
\textbf{\modelname} & 5/5            & 5/5            & 5/5            & 5/5            & 5/5            & \textbf{100\%}
\end{tabular}
\label{table:gum_result}
\end{table}

Fig. \ref{fig:result_gum_effective} displays some failure cases without \modelname. 
In these instances, slight prediction biases in the 2D image inputs result in significant 3D projection errors.
Such error leads to grasping failures, particularly when the robot approaches non-convex handles or views from a side angle. 
While \modelname$^*$ can resolve most of these cases, it fails once during testing. 
These results demonstrate the model's effectiveness in generalizing to diverse real-world scenarios.

\subsection{Ablation: Impact of feedback control}
\begin{table}[h]
\centering
\caption{Open vs. Closed loop}
\begin{tabular}{c|ccccc|c}
Method & door1   & door2  & door3  & door4 & door5 & sum    \\ \hline
Open-loop  & 3/5 & 1/5 & 2/5 & 1/5 & 3/5 & 40\% \\
\textbf{Closed-loop} & 5/5 & 5/5 & 5/5 & 5/5 & 5/5 & 100\% \\
\end{tabular}
\label{table:open_vs_closed}
\end{table}

We perform an ablation study to quantify the necessity of closed-loop feedback planning in open-world environments.
We compare our method with a simple open-loop implementation that predicts the primitives at the beginning and sequentially executes them.
When the robot tries to rotate or push/pull the door under this setting, we randomly sample a rotation direction or push/pull type.
In this open-loop implementation, the robot stops when it either succeeds or cannot proceed due to collision, failure to unlock, or other unexpected behavior during execution.

\begin{figure}[t]
\vspace{2mm}
\centering
\includegraphics[width=0.9\linewidth]{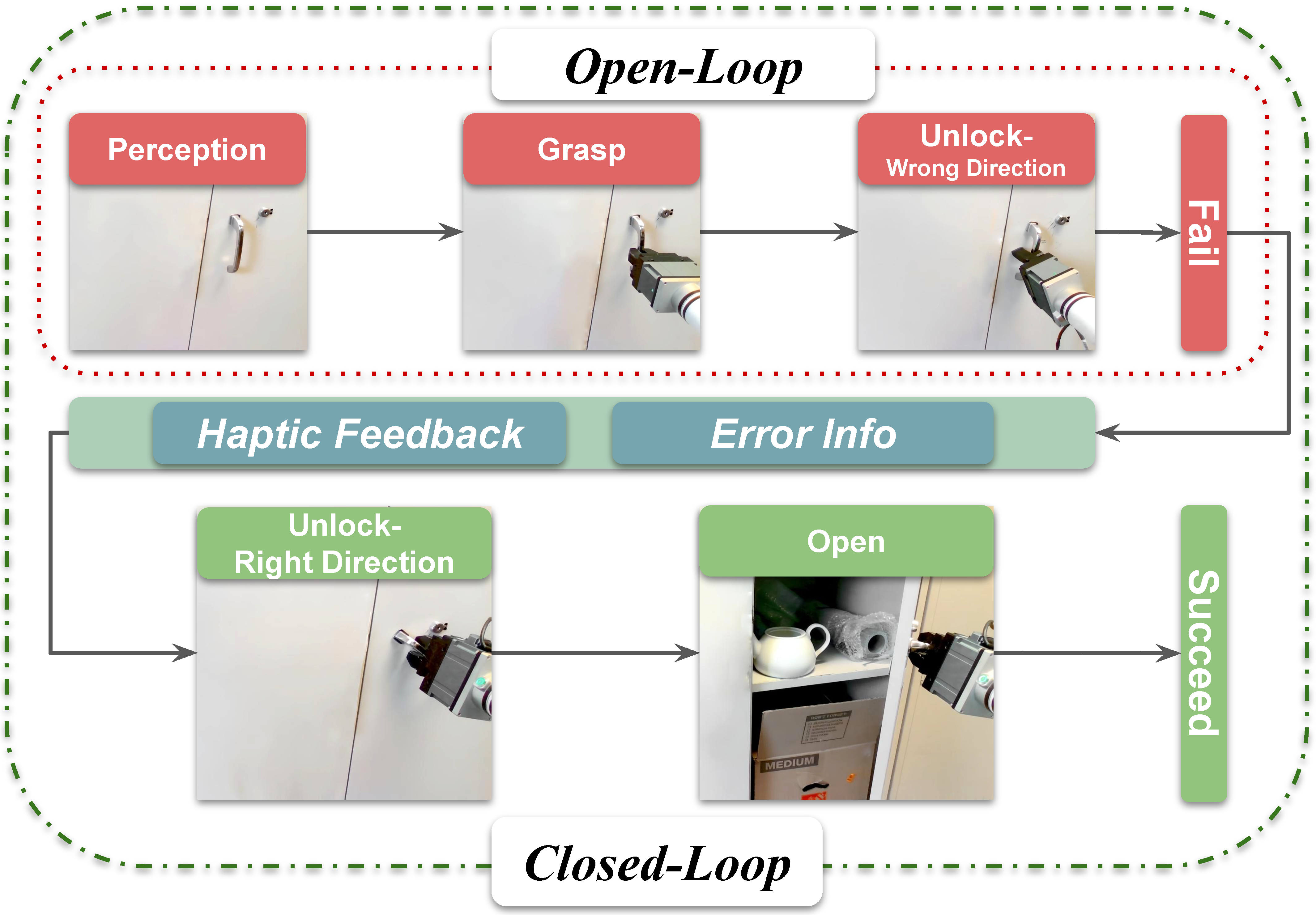}
\caption{With multi-modal feedback, our system can open the cabinet with an unknown unlocking direction via explore-and-adapt.}
\label{fig:result_closed_loop}
\vspace{-2mm}
\end{figure}

Table \ref{table:open_vs_closed} shows the quantitative results comparing performance with and without feedback planning on 5 randomly chosen doors.
Without closed-loop feedback, the robot has to guess the door's rotation direction and push/pull type. In contrast, the explore-and-adapt mechanism of our closed-loop architecture significantly improves success rates in such scenarios. 
Fig. \ref{fig:result_closed_loop} also illustrates how feedback prevents task failures in several cases.

\subsection{Ablation: Importance of haptic}
To explore whether the information from haptic feedback can be retrieved by visual input, we conduct an ablation study that replaces haptic feedback with predictions from VLMs.
In this experiment, we use the pre-trained CLIP ViT-B/32 model~\cite{radford2021learning} and Gemini 1.5 Pro~\cite{team2023gemini} to classify feedback types and the push/pull type of doors with solely the visual appearance of the door.
We then execute the corresponding action during the experiment, particularly in \textit{unlock} primitives and \textit{open}.

\begin{table}[h]
\centering
\caption{VLM vs. Haptic}
\begin{tabular}{c|cccc|c}
Method & grasp   & unlock-L  & unlock-K  & open  & push / pull  \\ \hline
CLIP & 42.67\% & 33.33\% & 38.81\% & 15.79\% & 15.00\% \\
Gemini& 29.63\% & 93.75\% & 86.36\% & 88.89\% & 65.00\% \\
\textbf{Haptics}  & 100\%   & 100\%   & 100\%   & 100\%  & 100\%
\end{tabular}
\label{table:vlm_vs_haptic}
\end{table}

Our results, shown in Table \ref{table:vlm_vs_haptic}, indicate that VLMs are not reliable for determining the state of motion primitives or door types. 
The haptic-based method significantly outperforms the vision-based models, further underscoring the importance of haptic feedback for reliably manipulating unseen doors.

\section{CONCLUSIONS}

We present a haptic-informed closed-loop control system for robust door-opening tasks in open environments. 
By integrating real-time haptic feedback, our framework enables mobile robots to dynamically adapt to non-visual properties during manipulation, improving generalization to unstructured, real-world scenarios.
Additionally, our visual perception module, trained on a small dataset of Internet images, enables accurate grasp pose detection, allowing effective generalization to various door types without extensive domain-specific training. 
Field tests show a success rate increase from 50\% to 90\%, demonstrating the system’s reliability in complex environments.
In future work, we plan to extend this framework to a wider range of manipulation tasks, incorporating learning-based policies for enhanced adaptability.



\bibliographystyle{IEEEtran}
\bibliography{ref}

\begin{thebibliography}{10}
\providecommand{\url}[1]{#1}
\csname url@samestyle\endcsname
\providecommand{\newblock}{\relax}
\providecommand{\bibinfo}[2]{#2}
\providecommand{\BIBentrySTDinterwordspacing}{\spaceskip=0pt\relax}
\providecommand{\BIBentryALTinterwordstretchfactor}{4}
\providecommand{\BIBentryALTinterwordspacing}{\spaceskip=\fontdimen2\font plus
\BIBentryALTinterwordstretchfactor\fontdimen3\font minus \fontdimen4\font\relax}
\providecommand{\BIBforeignlanguage}[2]{{%
\expandafter\ifx\csname l@#1\endcsname\relax
\typeout{** WARNING: IEEEtran.bst: No hyphenation pattern has been}%
\typeout{** loaded for the language `#1'. Using the pattern for}%
\typeout{** the default language instead.}%
\else
\language=\csname l@#1\endcsname
\fi
#2}}
\providecommand{\BIBdecl}{\relax}
\BIBdecl

\bibitem{6385835}
Y.~Karayiannidis, C.~Smith, F.~E. Viña, P.~Ogren, and D.~Kragic, ``“open sesame!” adaptive force/velocity control for opening unknown doors,'' in \emph{2012 IEEE/RSJ International Conference on Intelligent Robots and Systems}, 2012, pp. 4040--4047.

\bibitem{abraham2020model}
I.~Abraham, A.~Handa, N.~Ratliff, K.~Lowrey, T.~D. Murphey, and D.~Fox, ``Model-based generalization under parameter uncertainty using path integral control,'' \emph{IEEE Robotics and Automation Letters}, vol.~5, no.~2, pp. 2864--2871, 2020.

\bibitem{sturm2011probabilistic}
J.~Sturm, C.~Stachniss, and W.~Burgard, ``A probabilistic framework for learning kinematic models of articulated objects,'' \emph{Journal of Artificial Intelligence Research}, vol.~41, pp. 477--526, 2011.

\bibitem{kang2024versatile}
G.~Kang, H.~Seong, D.~Lee, and D.~H. Shim, ``A versatile door opening system with mobile manipulator through adaptive position-force control and reinforcement learning,'' \emph{Robotics and Autonomous Systems}, p. 104760, 2024.

\bibitem{gu2017deep}
S.~Gu, E.~Holly, T.~Lillicrap, and S.~Levine, ``Deep reinforcement learning for robotic manipulation with asynchronous off-policy updates,'' in \emph{2017 IEEE international conference on robotics and automation (ICRA)}.\hskip 1em plus 0.5em minus 0.4em\relax IEEE, 2017, pp. 3389--3396.

\bibitem{li2020hrl4in}
C.~Li, F.~Xia, R.~Martin-Martin, and S.~Savarese, ``Hrl4in: Hierarchical reinforcement learning for interactive navigation with mobile manipulators,'' in \emph{Conference on Robot Learning}.\hskip 1em plus 0.5em minus 0.4em\relax PMLR, 2020, pp. 603--616.

\bibitem{chebotar2017path}
Y.~Chebotar, M.~Kalakrishnan, A.~Yahya, A.~Li, S.~Schaal, and S.~Levine, ``Path integral guided policy search,'' in \emph{2017 IEEE international conference on robotics and automation (ICRA)}.\hskip 1em plus 0.5em minus 0.4em\relax IEEE, 2017, pp. 3381--3388.

\bibitem{ahn2022can}
M.~Ahn, A.~Brohan, N.~Brown, Y.~Chebotar, O.~Cortes, B.~David, C.~Finn, C.~Fu, K.~Gopalakrishnan, K.~Hausman \emph{et~al.}, ``Do as i can, not as i say: Grounding language in robotic affordances,'' \emph{arXiv preprint arXiv:2204.01691}, 2022.

\bibitem{liang2023code}
J.~Liang, W.~Huang, F.~Xia, P.~Xu, K.~Hausman, B.~Ichter, P.~Florence, and A.~Zeng, ``Code as policies: Language model programs for embodied control,'' in \emph{2023 IEEE International Conference on Robotics and Automation (ICRA)}.\hskip 1em plus 0.5em minus 0.4em\relax IEEE, 2023, pp. 9493--9500.

\bibitem{driess2023palm}
D.~Driess, F.~Xia, M.~S. Sajjadi, C.~Lynch, A.~Chowdhery, B.~Ichter, A.~Wahid, J.~Tompson, Q.~Vuong, T.~Yu \emph{et~al.}, ``Palm-e: An embodied multimodal language model,'' \emph{arXiv preprint arXiv:2303.03378}, 2023.

\bibitem{brohan2023rt}
A.~Brohan, N.~Brown, J.~Carbajal, Y.~Chebotar, X.~Chen, K.~Choromanski, T.~Ding, D.~Driess, A.~Dubey, C.~Finn \emph{et~al.}, ``Rt-2: Vision-language-action models transfer web knowledge to robotic control,'' \emph{arXiv preprint arXiv:2307.15818}, 2023.

\bibitem{arduengo2021robust}
M.~Arduengo, C.~Torras, and L.~Sentis, ``Robust and adaptive door operation with a mobile robot,'' \emph{Intelligent Service Robotics}, vol.~14, no.~3, pp. 409--425, 2021.

\bibitem{stuede2019door}
M.~Stuede, K.~Nuelle, S.~Tappe, and T.~Ortmaier, ``Door opening and traversal with an industrial cartesian impedance controlled mobile robot,'' in \emph{2019 International Conference on Robotics and Automation (ICRA)}.\hskip 1em plus 0.5em minus 0.4em\relax IEEE, 2019, pp. 966--972.

\bibitem{5649847}
E.~Klingbeil, A.~Saxena, and A.~Y. Ng, ``Learning to open new doors,'' in \emph{2010 IEEE/RSJ International Conference on Intelligent Robots and Systems}, 2010, pp. 2751--2757.

\bibitem{tremblay2018deep}
J.~Tremblay, T.~To, B.~Sundaralingam, Y.~Xiang, D.~Fox, and S.~Birchfield, ``Deep object pose estimation for semantic robotic grasping of household objects,'' \emph{arXiv preprint arXiv:1809.10790}, 2018.

\bibitem{chu2019learning}
F.-J. Chu, R.~Xu, and P.~A. Vela, ``Learning affordance segmentation for real-world robotic manipulation via synthetic images,'' \emph{IEEE Robotics and Automation Letters}, vol.~4, no.~2, pp. 1140--1147, 2019.

\bibitem{9206039}
J.~Wang, S.~Lin, C.~Hu, Y.~Zhu, and L.~Zhu, ``Learning semantic keypoint representations for door opening manipulation,'' \emph{IEEE Robotics and Automation Letters}, vol.~5, no.~4, pp. 6980--6987, 2020.

\bibitem{qin2023dexpoint}
Y.~Qin, B.~Huang, Z.-H. Yin, H.~Su, and X.~Wang, ``Dexpoint: Generalizable point cloud reinforcement learning for sim-to-real dexterous manipulation,'' in \emph{Conference on Robot Learning}.\hskip 1em plus 0.5em minus 0.4em\relax PMLR, 2023, pp. 594--605.

\bibitem{urakami2019doorgym}
Y.~Urakami, A.~Hodgkinson, C.~Carlin, R.~Leu, L.~Rigazio, and P.~Abbeel, ``Doorgym: A scalable door opening environment and baseline agent,'' \emph{arXiv preprint arXiv:1908.01887}, 2019.

\bibitem{xiong2024adaptive}
H.~Xiong, R.~Mendonca, K.~Shaw, and D.~Pathak, ``Adaptive mobile manipulation for articulated objects in the open world,'' \emph{arXiv preprint arXiv:2401.14403}, 2024.

\bibitem{ding2021sim}
Z.~Ding, Y.-Y. Tsai, W.~W. Lee, and B.~Huang, ``Sim-to-real transfer for robotic manipulation with tactile sensory,'' in \emph{2021 IEEE/RSJ International Conference on Intelligent Robots and Systems (IROS)}.\hskip 1em plus 0.5em minus 0.4em\relax IEEE, 2021, pp. 6778--6785.

\bibitem{yuan2017gelsight}
W.~Yuan, S.~Dong, and E.~H. Adelson, ``Gelsight: High-resolution robot tactile sensors for estimating geometry and force,'' \emph{Sensors}, vol.~17, no.~12, p. 2762, 2017.

\bibitem{calandra2018more}
R.~Calandra, A.~Owens, D.~Jayaraman, J.~Lin, W.~Yuan, J.~Malik, E.~H. Adelson, and S.~Levine, ``More than a feeling: Learning to grasp and regrasp using vision and touch,'' \emph{IEEE Robotics and Automation Letters}, vol.~3, no.~4, pp. 3300--3307, 2018.

\bibitem{karayiannidis2016adaptive}
Y.~Karayiannidis, C.~Smith, F.~E.~V. Barrientos, P.~{\"O}gren, and D.~Kragic, ``An adaptive control approach for opening doors and drawers under uncertainties,'' \emph{IEEE Transactions on Robotics}, vol.~32, no.~1, pp. 161--175, 2016.

\bibitem{jain2008behaviors}
A.~Jain and C.~C. Kemp, ``Behaviors for robust door opening and doorway traversal with a force-sensing mobile manipulator,'' in \emph{RSS Manipulation Workshop: Intelligence in Human Environments}.\hskip 1em plus 0.5em minus 0.4em\relax Citeseer, 2008.

\bibitem{van2015learning}
H.~Van~Hoof, T.~Hermans, G.~Neumann, and J.~Peters, ``Learning robot in-hand manipulation with tactile features,'' in \emph{2015 IEEE-RAS 15th International Conference on Humanoid Robots (Humanoids)}.\hskip 1em plus 0.5em minus 0.4em\relax IEEE, 2015, pp. 121--127.

\bibitem{xu2022towards}
H.~Xu, Y.~Luo, S.~Wang, T.~Darrell, and R.~Calandra, ``Towards learning to play piano with dexterous hands and touch,'' in \emph{2022 IEEE/RSJ International Conference on Intelligent Robots and Systems (IROS)}.\hskip 1em plus 0.5em minus 0.4em\relax IEEE, 2022, pp. 10\,410--10\,416.

\bibitem{fischler1981random}
M.~A. Fischler and R.~C. Bolles, ``Random sample consensus: a paradigm for model fitting with applications to image analysis and automated cartography,'' \emph{Communications of the ACM}, vol.~24, no.~6, pp. 381--395, 1981.

\bibitem{zhou2022detecting}
X.~Zhou, R.~Girdhar, A.~Joulin, P.~Kr{\"a}henb{\"u}hl, and I.~Misra, ``Detecting twenty-thousand classes using image-level supervision,'' in \emph{ECCV}, 2022.

\bibitem{kirillov2023segment}
A.~Kirillov, E.~Mintun, N.~Ravi, H.~Mao, C.~Rolland, L.~Gustafson, T.~Xiao, S.~Whitehead, A.~C. Berg, W.-Y. Lo \emph{et~al.}, ``Segment anything,'' in \emph{Proceedings of the IEEE/CVF International Conference on Computer Vision}, 2023, pp. 4015--4026.

\bibitem{RealMan2024CompoundRobot}
{RealMan Robotics}, ``{Compound Robot - RealMan Robotics},'' \url{https://www.realman-robotics.com/fuhejiqiren.html}, 2024, accessed: 2024-09-07.

\bibitem{DHRobotics2024PGI}
{DH-Robotics}, ``{PGI Series Industrial Electric Parallel Gripper},'' \url{https://en.dh-robotics.com/product/pgi}, 2024, accessed: 2024-09-07.

\bibitem{team2023gemini}
G.~Team, R.~Anil, S.~Borgeaud, Y.~Wu, J.-B. Alayrac, J.~Yu, R.~Soricut, J.~Schalkwyk, A.~M. Dai, A.~Hauth \emph{et~al.}, ``Gemini: a family of highly capable multimodal models,'' \emph{arXiv preprint arXiv:2312.11805}, 2023.

\bibitem{radford2021learning}
A.~Radford, J.~W. Kim, C.~Hallacy, A.~Ramesh, G.~Goh, S.~Agarwal, G.~Sastry, A.~Askell, P.~Mishkin, J.~Clark \emph{et~al.}, ``Learning transferable visual models from natural language supervision,'' in \emph{International conference on machine learning}.\hskip 1em plus 0.5em minus 0.4em\relax PMLR, 2021, pp. 8748--8763.

\end{thebibliography}

\end{document}